%% file: bmvc_final.tex
\documentclass{bmvc2k}

\input{preamble}

\title{SignMatch: Matching Dictionary Signs to Continuous Sign Language Video}

\addauthor{Ryan Wong}{https://ryanwongsa.github.io/}{1}
\addauthor{Youngjoon Jang}{https://art-jang.github.io/}{1}
\addauthor{Liliane Momeni}{https://www.robots.ox.ac.uk/~liliane/}{1}
\addauthor{G\"ul Varol}{http://imagine.enpc.fr/~varolg}{1,2}
\addauthor{Andrew Zisserman}{https://robots.ox.ac.uk/~az}{1}

\addinstitution{
 Visual Geometry Group\\
 Department of Engineering Science\\
 University of Oxford\\
 UK
}
\addinstitution{
 LIGM, École des Ponts, IP Paris, Univ Gustave Eiffel, CNRS\\
 France
}

\runninghead{Wong Et Al.}{SignMatch}

\begin{document}

\maketitle
\input{secs/0_abs}

\input{secs/1_intro}

\input{secs/2_related}

\input{secs/3_method}

\input{secs/4_training}

\input{secs/5_exps}

\input{secs/6_conclusion}

\subsubsection*{Acknowledgements}

The BOBSL images in this paper are used with the kind permission of the BBC.
This work was supported by the UKRI EPSRC Programme Grant SignGPT EP/Z535370/1, and a Royal Society
Research Professorship RSRP$\backslash$R$\backslash$241003. The authors acknowledge the use of resources provided by the Isambard-AI National AI Research Resource (AIRR) \cite{mcintosh2024isambard}. Isambard-AI is operated by the University of Bristol and is funded by the UK Government’s Department for Business, Innovation, Science and Trade (DBIST) via UK Research and Innovation; and the Science and Technology Facilities Council [ST/AIRR/I-A-I/1023].

\bibliography{shortstrings, main}

% \bibliography{egbib}

\newpage

\input{secs/7_appendices}

\end{document}

%% file: preamble.tex
\usepackage{booktabs}       % professional-quality tables
\usepackage{amssymb}
\usepackage{wrapfig}
\usepackage{cleveref}
\usepackage{xcolor}
\usepackage{colortbl}
\definecolor{rowline}{gray}{0.85}
\usepackage[normalem]{ulem}
\providecommand{\drsh}{\ensuremath{\hookrightarrow}}

\newcommand{\newpara}[1]{\vspace{3pt}\noindent\textbf{#1}}

\crefname{equation}{Eq.}{Eqs.}
\Crefname{equation}{Equation}{Equations}

\crefname{figure}{Fig.}{Figs.}
\Crefname{figure}{Figure}{Figures}

\crefname{table}{Tab.}{Tabs.}
\Crefname{table}{Table}{Tables}

\crefname{section}{Sec.}{Secs.}
\Crefname{section}{Section}{Sections}

\crefname{algorithm}{Alg.}{Algs.}
\Crefname{algorithm}{Algorithm}{Algorithms}

\definecolor{matchgreen}{rgb}{0.10,0.45,0.18}  \definecolor{missred}{rgb}{0.70,0.12,0.12}   \newcommand{\ok}[1]{\textcolor{matchgreen}{#1}}
\newcommand{\miss}[1]{\textcolor{missred}{#1}}

\newcommand{\rowsep}{\arrayrulecolor{rowline}\hline\arrayrulecolor{black}}

%% file: secs/0_abs.tex
\begin{abstract}

The objective of this paper is to match dictionary sign videos to corresponding signs in continuous signing videos, where a match is defined by the visual similarity alone -- the handshape and motion relative to the body. To achieve this, we learn a prototype-structured sign embedding space from continuous video annotated with signs, where each learnable prototype corresponds to a sign class.  Isolated dictionary videos are then mapped into this sign space, enabling the matching between dictionary exemplars and continuous sign instances. This design supports direct dictionary-guided sign matching through embedding similarity and naturally extends to unseen signs using only dictionary exemplars. 
Experiments on ASL-Citizen dictionary retrieval, ChaLearn OSLWL dictionary-to-continuous sign matching, and using BOBSL's CSLR2 evaluation for automatic sign annotation demonstrate strong generalisation across datasets, tasks and sign languages. Without benchmark-specific supervision, the learned representation transfers effectively across American, British, and Spanish Sign Languages, outperforming prior methods on all three benchmarks. Project page: \url{https://www.robots.ox.ac.uk/~vgg/research/signmatch/}

\end{abstract}

%% file: secs/1_intro.tex
\section{Introduction}
\begin{figure}[t]
    \centering
    \includegraphics[width=1.0\linewidth]{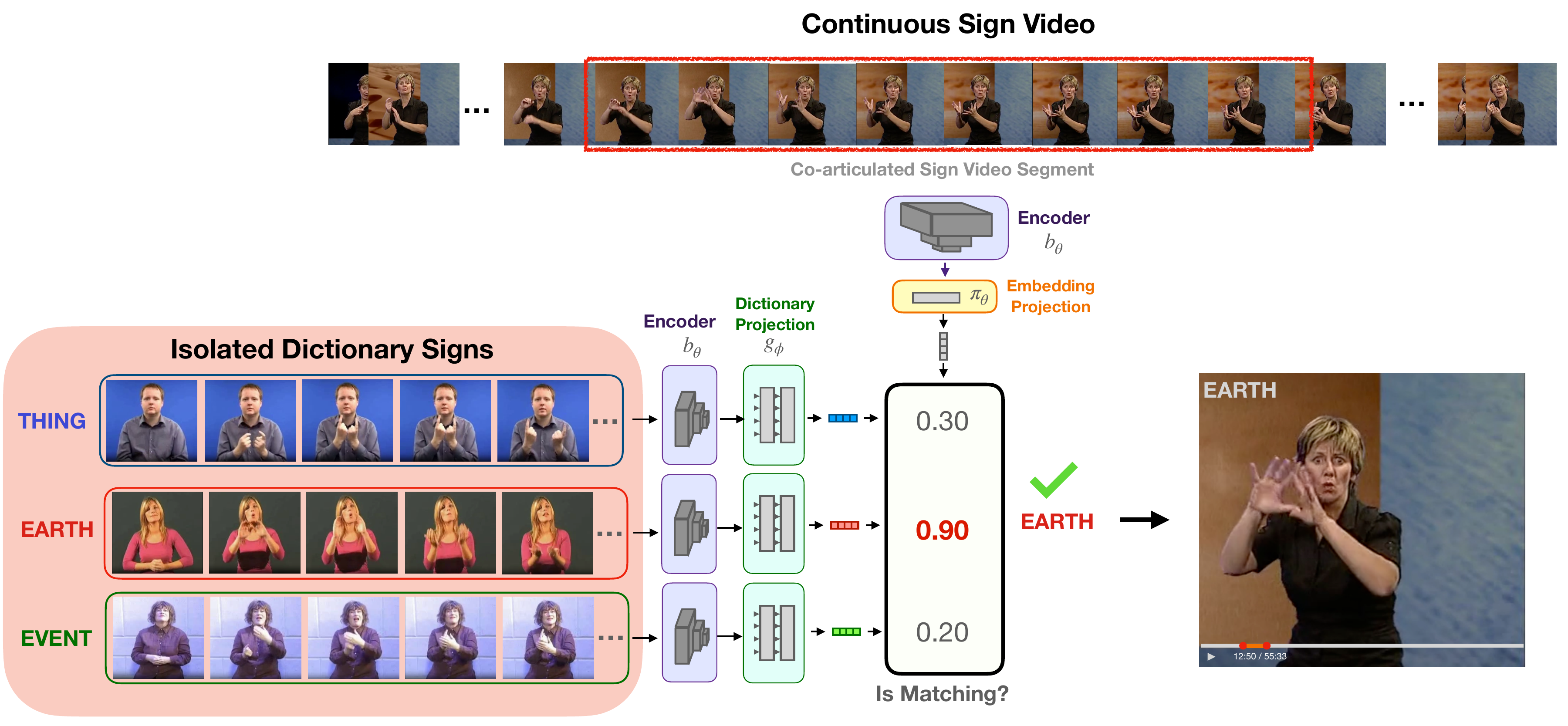}
    \caption{{\bf Sign annotation using a matching network.} Unseen dictionary 
sign videos are enrolled by generating their embeddings. The continuous signing video can then be labelled
with these new signs (if they are present) at the location where they occur simply by computing a cosine similarity score.}
    \label{fig:teaser}
    % \vspace{-0.4cm}
\end{figure}

Sign languages are natural visual languages used by Deaf communities and consist of complex manual and non-manual articulations performed over time~\cite{sutton1999linguistics}. 
The objective of this paper is to visually match signs between dictionaries and continuous signing  -- 
given a sequence of continuous signing and a set of dictionary video sign examples, the goal is to
identify and localise temporal segments in the continuous video that {\em visually} correspond to the dictionary signs. 

Having the ability to visually match signs enables multiple applications: first, and most directly, it provides
a means to look up in the dictionary the meaning of signs in the continuous video; second, 
it enables scalable annotation of sign language datasets for
signs using only isolated dictionary examples; third, it enables search by sign of signing videos; and finally, 
by transitivity, it enables translations between sign dictionaries.

Visual sign matching depends only on the visual realisation, i.e.\ on
the sign `production', in particular on the handshape and hand movement
relative to the body. It does not depend on the semantics or meaning
of the sign. The challenge is that dictionary examples are {\em isolated} in
the sense that  the signer typically starts from a rest position,
performs the sign, and then returns to the rest position, a production known as
citation form~\cite{neidle2022documentation}. 
In
contrast, in continuous signing, e.g.\ a conversation or
interpretation, signs vary significantly due to contextual
co-articulation from the signs occurring before and after, as well as
variation arising from signer motion, appearance, and execution speed -- which is often considerably faster than that
of the dictionary production.

Existing approaches to sign matching
with dictionaries~\cite{momeni2020watch} typically learn a shared embedding space
between isolated dictionary signs and co-articulated signing. 
However, constructing such a space is not well constrained because of the substantial differences between
these two domains, and also due to the fact that dictionaries are {\em sparse} and homogeneous, often containing only one or a few examples per sign despite substantial intra-class variation, and performed by a small number of individuals.  In contrast, continuous signing video datasets
can contain multiple different signers and have multiple instances of each sign,  and are virtually unlimited in scale.

The key idea in this paper is to 
learn the sign embedding space, 
not directly from the sparse dictionary exemplars,
but from temporal annotations of multiple sign instances and signers
in co-articulated continuous video sequences. This can be collected at
far larger scale than curated dictionaries, and exposes natural
variation across signers, execution speeds, and contextual
co-articulation, providing far richer supervision than isolated dictionary
exemplars. We organise this space around signs as {\em prototypes}, and map
each dictionary example into the same space through its embedding,
yielding a single space in which both domains are directly comparable. Overall, this approach
decouples representation learning from dictionary alignment.
\Cref{fig:teaser} illustrates how dictionary examples can be used to annotate continuous signing, using
a matching network built on this embedding space.

A crucial question is whether this approach generalises beyond the
signs it has been trained on. We show that indeed it does
generalise both to unseen signs from the languages it has been trained
on {\em and} also to new sign languages entirely.  This is a step
towards a {\em universal} sign embedding, able to represent
the production of the sign, but agnostic about the language (American,
British, Spanish \ldots) and semantics. 

In summary, we make the following three contributions: (1) we introduce an approach that uses temporally annotated continuous signing to learn an embedding space using learnable sign prototypes; (2) we learn a mapping from isolated dictionary videos into this embedding space, thereby enabling continuous sign-matching with dictionaries; and (3) we demonstrate that the learned representation generalises across tasks, datasets, and sign languages: on ASL-Citizen it transfers to dictionary retrieval for unseen signs,  without dataset-specific training; on the ChaLearn OSLWL benchmark it enables dictionary to continuous sign matching in Spanish Sign Language despite training on BSL/ASL data; and on BOBSL it supports large-scale automatic annotation through dictionary-guided retrieval. In {\em all} cases the performance on sign matching surpasses the performance of previous methods on these public benchmarks.

%% file: secs/2_related.tex
\section{Related Work}
\paragraph{Sign spotting from continuous signing.}
Sign spotting aims to localise occurrences of a target sign within continuous, co-articulated video. Early approaches combined hand-crafted features with temporal matching methods such as dynamic time warping and hierarchical sequential patterns~\cite{ong2014sign,viitaniemi2014s,yang2008sign,yang2006detecting,Santemiz2009AutomaticSS}, whereas recent work learns deep spatio-temporal representations from larger-scale video, typically supervised by weakly aligned subtitles~\cite{albanie2020bsl,momeni2020watch,li2020transferring}. The task itself has been framed in different ways: the ECCV 2022 Sign Spotting Challenge~\cite{vazquez2022eccv} distinguished a multiple-shot supervised setting (MSSL), which resembles continuous recognition over a fixed vocabulary, from a one-shot setting (OSLWL) that matches isolated dictionary queries against candidate intervals in continuous video. Since dense manual annotation does not scale, a central question across these settings is where the supervision comes from.

\newpara{Mouthing-based supervision for sign spotting.} 
A major scalable source of supervision is {\em mouthing}: signers often mouth the corresponding spoken words while signing, and these cues have been used both to improve sign language translation and recognition~\cite{wu2025signmouth,jang2025lost,thomas2025signbind,shi2022open,koller2014read,koller2015deep} and to mine sign annotations at scale. BSL-1K~\cite{albanie2020bsl} introduced a pipeline that combines weakly aligned subtitles with visual keyword spotting over mouth movements~\cite{momeni2020seeing,stafylakis2018zero} to recover sparse sign labels from broadcast footage, enabling large-scale training of co-articulated recognition models. However, mouthing supervision is inherently limited: not every sign is mouthed, the labels are tied to spoken-language words, and annotations are sparse in time. It therefore captures only one channel of the signal and largely ignores the manual articulation that defines a sign, motivating supervision anchored in the sign itself rather than in spoken language.

\newpara{Dictionary-based sign spotting.}
Dictionary-based sign spotting uses isolated dictionary exemplars as visual queries to localise signs in continuous video, offering broad lexical coverage beyond the signs annotated in continuous corpora. The difficulty is a domain gap: dictionary signs are isolated, canonical, and slowly articulated, whereas continuous signs are shorter and shaped by co-articulation, context, and signer variation. \cite{momeni2020watch,varol2022scaling} bridge this gap by learning cross-domain embeddings with a multiple-instance Noise Contrastive Estimation (NCE) objective, constructing bags of candidate positive and negative matches because no frame-level dictionary--continuous correspondences are available. This is an effective use of weak supervision, but it couples representation learning, domain alignment, subtitle disambiguation, dictionary-variant selection, and negative sampling into a single objective. 
One-shot matchers instead align an isolated query directly to a target clip~\cite{jiang2021looking}; more broadly, this connects to low-shot temporal localisation, where a segment in an untrimmed video is matched to one or more query clips~\cite{feng2018video,yang2018one,cao2020few}.
In contrast, our approach decouples representation learning from sparse dictionary alignment: it first learns a reusable embedding from co-articulated data and then enrols dictionary exemplars into that space.

\newpara{Dictionary retrieval and sign-centric lookup.}
A related line of work lets users retrieve dictionary entries by performing a sign rather than knowing its written translation. GlossFinder~\cite{xu2022automatic} presents a webcam-based dictionary interface that returns candidate glosses from a user's signing, and ASL Citizen~\cite{desai2023asl} formalises isolated sign recognition as dictionary retrieval, with a large community-sourced dataset evaluated by retrieval metrics such as recall-at-$K$. These methods address isolated-to-isolated retrieval, whereas we use isolated dictionary signs to retrieve and localise instances within continuous co-articulated signing.

\newpara{Sign language corpora and automatic annotations.}
Progress in sign spotting depends heavily on continuous sign language corpora, whose supervision is usually only weakly aligned through subtitles or similar cues~\cite{Albanie2021bobsl,uthus2023youtube}. \Cref{tab:corpora} lists the corpora used in this work and illustrates the field's shift from small, carefully annotated resources to large broadcast- and web-scale collections. Early linguistically motivated datasets such as the BSL Corpus~\cite{schembri2013building} gave way to broadcast corpora mined with mouthing and subtitle cues (BSL-1K~\cite{albanie2020bsl}, BOBSL~\cite{Albanie2021bobsl}) and, more recently, to open-domain video supervised by captions (YouTube-ASL~\cite{uthus2023youtube}, YouTube-SL-25~\cite{tanzer2025youtube}). As corpora grow but dense manual labels remain scarce, automatic retrieval and annotation become increasingly important for downstream recognition~\cite{ahn2024slowfast,jang2023self,zheng2023cvt,li2020word}, alignment~\cite{jang2026deep,bull2021aligning,jiang2025segment}, retrieval~\cite{raude2024}, and translation~\cite{jang2025lost,li2025uni,wong2024sign2gpt,tanzer2024fleurs}.

\begin{table}[t]
    \centering
    \setlength{\tabcolsep}{6pt}
    \resizebox{\linewidth}{!}
    {
        \begin{tabular}{lll}
        \toprule
        Corpus & Scale (hours) & Participants / signers \\
        \midrule
        BSL Corpus~\cite{schembri2013building} & 125 (partially annotated) & 249 participants \\
        BOBSL~\cite{Albanie2021bobsl} & 1,447 & 39 signers \\
        YouTube-ASL~\cite{uthus2023youtube} & 984 & $>$2,519 signers \\
        YouTube-SL-25~\cite{tanzer2025youtube} & 3,207 across $>$25 sign languages & $>$3,072 channels (approx.\ lower bound on signers) \\
        YouTube-SL-25 (BSL subset) & 74 & 60 channels \\
        \bottomrule
        \end{tabular}
    }
    \caption{Overview of the continuous sign language corpora.}
    \label{tab:corpora}
\end{table}

%% file: secs/3_method.tex
\section{The Sign Matching Model}

\subsection{Method Overview}

Our goal is to determine whether an isolated dictionary sign video and a co-articulated signing segment visually correspond to the same sign, as illustrated in \Cref{fig:teaser}. In this section, we describe a matching network that can robustly determine whether the two videos match. 

\paragraph{Training.} 
We assume access to co-articulated sign segments labelled with their corresponding dictionary entries for a subset of signs. From this paired supervision we learn a matching model that generalises to previously unseen dictionary data, in two stages: (i) an initialisation stage that learns a structured representation space from co-articulated signing using prototypes, and (ii) a dictionary alignment stage that maps isolated dictionary signs   into this space. This division reflects the differences between the two domain. Dictionaries contain only one or a few examples per sign, whereas co-articulated signing data provide many instances across signers and capture variation in speed and context that citation-form exemplars lack. We therefore learn the representation geometry from co-articulated signing and subsequently align dictionary exemplars to it. In addition, citation-form signs are typically articulated more slowly and span more frames, so dictionary videos are windowed and aggregated rather than matched at a fixed length.
We present these stages in the following sections. 

\newpara{Inference.}
\label{sec:dict_spotting_in_coart}
At inference time, embeddings of an isolated dictionary sign are matched against embeddings of the continuous signing video to determine whether the dictionary sign is present, as illustrated in \Cref{fig:teaser}. Dictionary videos are represented as $x^d_{1:T}$ and encoded by a backbone followed by a temporal aggregation head to obtain a dictionary embedding, $e^d = g_\phi(b_\theta(x^d_{1:T}))$. Co-articulated video segments $x$ are encoded by the same backbone and a projection head to obtain segment embeddings, $e = \pi_\theta(b_\theta(x))$. A high cosine similarity score between the embeddings indicates likely occurrences of the queried dictionary sign in the corresponding co-articulated segment. 

\begin{figure}[t]
    \centering
    \includegraphics[width=1.0\linewidth]{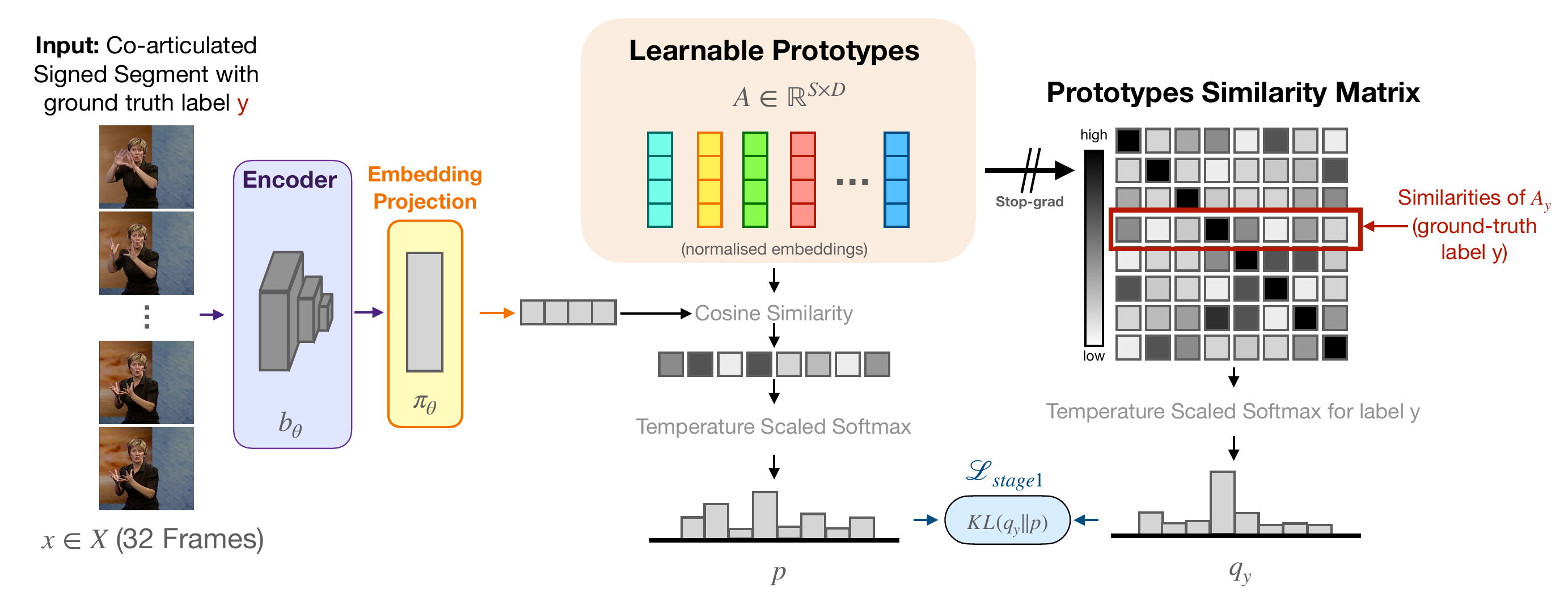}
    \caption{
    Stage~1 learns the sign embedding space from co-articulated signing data. Given an input segment $x$ with ground-truth label $y$, a backbone $b_\theta$ and projection head $\pi_\theta$ produce an embedding compared against a set of learnable sign prototypes ${A_k}$ yielding a predicted class distribution $p$. Independently, similarities between the prototype corresponding to the ground-truth label $y$ and all prototypes define a soft target distribution $q_y$. The bar plots illustrate these predicted and target distributions. Training minimises the KL divergence between $q_y$ and $p$, encouraging examples of class y to assign probability mass to prototypes that are similar to $A_y$. As a result, the embedding space is organised according to prototype similarity, bringing visually similar sign classes closer together while separating dissimilar classes.
    }
    \label{fig:stage1}
\end{figure}

\subsection{Stage 1: Learning the sign embedding space using prototypes}

The goal of Stage 1 is to learn a structured embedding space from large-scale co-articulated signing data temporally annotated with sign classes $\mathcal S$, where the structure is induced by a set of learnable sign prototypes.

As illustrated in \Cref{fig:stage1}, a co-articulated sign segment is first encoded into an embedding representation. Let $b_\theta$ denote the video backbone and $\pi_\theta$ the projection head. Given a co-articulated window $x$ with label $y \in \mathcal S$, we obtain the embedding

\begin{equation}
e = \pi_\theta(b_\theta(x)) \in \mathbb{R}^{D}.
\end{equation}

We associate each sign class $k \in \mathcal S$  with a learnable prototype $A_k \in \mathbb{R}^{D}$, where $D$ is the embedding dimension. These prototypes represent canonical locations of sign classes within the embedding space. Since some sign classes may correspond to the same underlying
production, we do not force prototypes to be strictly distinct.
However, we would like visually similar signs to occupy nearby regions of the embedding space while remaining discriminative. To achieve this, we  use the relationships between the prototypes to define a soft supervision signal that is used to learn the backbone $b_\theta$,
projection head $\pi_\theta$, and the prototypes $A_k$.

To determine how the embedding $e$ of the co-articulated sign segment relates to the learned prototype space, we measure its similarity to each prototype.
We compute the cosine similarity between the normalised embedding ($\bar e$) and each normalised class prototype ($\bar A_k$):
\begin{equation}
z_k = \langle \bar e, \bar A_k \rangle,
\qquad k \in \mathcal S.
\label{eq_zk}
\end{equation}

The resulting similarity scores are converted into a predicted class distribution $p(\cdot \mid x)$ using a temperature-scaled softmax with temperature $\tau$, where $\tau$ controls the sharpness of the distribution.

For a training example with label $y$, we compute the cosine similarity between the corresponding prototype and all prototypes:
\begin{equation}
m_k = \langle \bar A_y, \bar A_k \rangle,
\qquad k \in \mathcal S.
\end{equation}

The resulting scores are converted into a target distribution $q_y$ using the same temperature-scaled softmax, yielding a soft target distribution over the prototype set. Since $\langle \bar A_y, \bar A_y \rangle = 1$, the target distribution assigns the highest mass to the class associated with the input while still allocating non-zero mass to visually related classes.

We then minimise the KL divergence between the prototype-induced target distribution ($q_y$) and the predicted class distribution, encouraging the model to match the similarity structure encoded by the prototypes:
\begin{equation}
\mathcal L_{\mathrm{stage1}}
=
\mathrm{KL}
\big(
q_y
\;\|\;
p(\cdot \mid x)
\big).
\end{equation}

The target distribution $q_y$ is detached during optimisation such that gradients are not propagated through the target branch. This prevents trivial self-reinforcement of the prototype similarities and stabilises learning of the prototype geometry. Under the KL objective, the video backbone $b_\theta$, projection head $\pi_\theta$, and prototype set ${A_k}$ are jointly optimised through the shared prototype space, encouraging sample embeddings to reproduce the similarity structure induced by the prototypes. As a result, visually related sign classes occupy nearby regions of the embedding space while remaining discriminative, thereby defining the reference geometry used to align isolated dictionary signs in Stage~2.

\subsection{Stage 2: Dictionary Alignment}

\begin{figure}[t]
    \centering
    \includegraphics[width=1.0\linewidth]{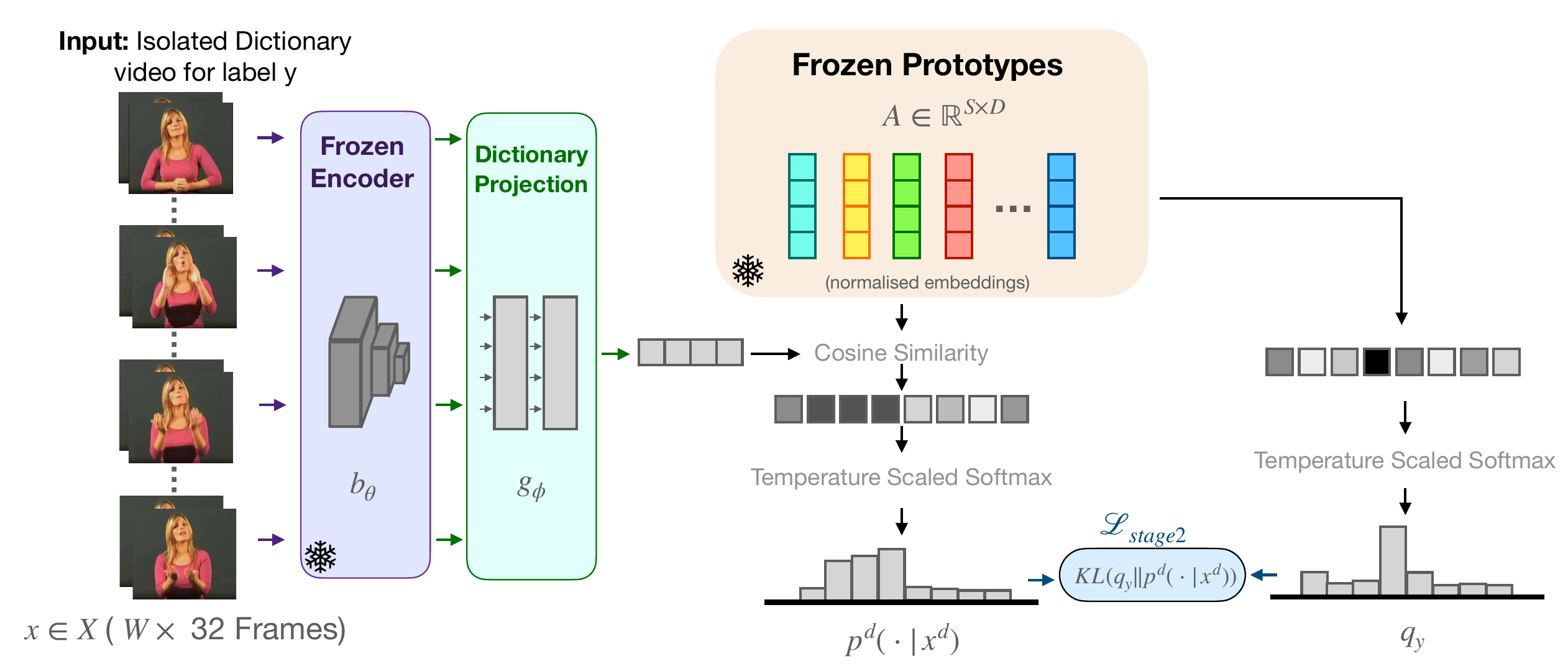}
    \caption{Stage 2 aligns isolated dictionary signs to the prototype space learned in Stage~1. A dictionary video is encoded using the frozen backbone $b_\theta$, and a dictionary projection network  $g_\phi$ aggregates features extracted from multiple temporal clips to produce a dictionary embedding. The embedding is compared against the frozen prototype set to obtain a predicted distribution $p^d(\cdot|x^d)$, while the prototype corresponding to the ground-truth label defines the target distribution $q_y$. The bar plots illustrate these predicted and target distributions. Training minimises the KL divergence between $q_y$ and $p^d(\cdot|x^d)$, learning a mapping from isolated dictionary signs into the embedding space learned from co-articulated signing.}
    \label{fig:stage2}
\end{figure}

The goal of Stage 2 is to align isolated dictionary signs to the embedding space learned in Stage 1. Since isolated dictionary videos provide only a single or small number of examples per sign, they are insufficient to learn a robust embedding space from scratch. Instead, we preserve the geometry learned from co-articulated signing and learn a mapping from dictionary videos into this space.

As illustrated in \Cref{fig:stage2}, a dictionary video is first encoded using the backbone learned in Stage~1. We freeze both the backbone $b_\theta$ and the learned prototype set $\{A_k\}$, and optimise only a dictionary projection network $g_\phi$. This ensures that the geometry learned from co-articulated signing remains fixed, and only the mapping from dictionary videos into this space is learned.

Dictionary signs are typically performed in isolation and often span more frames than co-articulated sign segments. To capture the full articulation of the sign, we extract $W$ overlapping clips of 32 frames using a sliding window with stride 16 and compute frozen backbone features for each clip.
\begin{equation}
h_i^d = b_\theta(x_i^d),
\qquad
i = 1,\dots,W.
\end{equation}

The dictionary projection network then aggregates these features to produce a dictionary embedding:
\begin{equation}
e^d = g_\phi(h_{1:W}^d) \in \mathbb{R}^{D}.
\end{equation}

To determine how the dictionary sign relates to the learned prototype space, we compare the dictionary embedding against all frozen prototypes. Using the same prototype-matching formulation as Stage~1, we compute a predicted distribution $p^d(\cdot \mid x^d)$ over the prototype set. The target distribution $q_y$ is computed using the frozen prototype geometry exactly as in Stage~1, preserving the inter-class relationships learned from co-articulated signing.

The Stage~2 objective is therefore

\begin{equation}
\mathcal L_{\mathrm{stage2}}
=
\mathrm{KL}
\big(
q_y
\;\|\;
p^d(\cdot \mid x^d)
\big).
\end{equation}

By matching the dictionary-induced distribution to the prototype-induced target distribution, the dictionary projection network learns to map isolated dictionary signs into the structured embedding space learned from co-articulated signing while preserving the inter-class relationships encoded by the frozen prototypes.

Once Stage 1 and Stage 2 are trained, the models can be used as described in Section~\ref{sec:dict_spotting_in_coart} to find matches of dictionary videos to their corresponding localisations in continuous sign videos.

%% file: secs/4_training.tex
\section{Training details}

This section describes the dictionary resources used for alignment, the model architecture, the two-stage training procedure, and the pseudo-label generation and training strategy used to scale supervision to larger sign language corpora.

\newpara{Sign language resources.}
The continuous sign language corpora used in this work were introduced in \Cref{tab:corpora}. Among these datasets, only BSLCorpus provides manual \textit{id-gloss} annotations, where each sign occurrence is assigned a unique lexical identifier that can be linked to a dictionary entry in BSL SignBank.
To extend supervision beyond BSLCorpus and enable training on additional corpora and sign languages, we incorporate external dictionary resources.

For BSL, we use BSL SignBank \cite{cormier2012corpus} and BSLDict \cite{momeni2020watch}. To support training and evaluation on ASL, we collect ASLDict, a large-scale dictionary resource derived from SignASL (\url{www.signasl.org}). An overview of the dictionary resources used in this work is provided in \Cref{tab:dictionaries}. Similar to BSLDict, ASLDict contains isolated sign videos rather than co-articulated signing.

\begin{table}[th]
    \centering
    \setlength{\tabcolsep}{20pt}
    \resizebox{\linewidth}{!}
    {
        \begin{tabular}{lll}
        \toprule
        Resource & Language & Scale \\
        \midrule
        BSL SignBank \cite{cormier2012corpus} & BSL & $\sim$2K sign entries \\
        BSLDict \cite{momeni2020watch} & BSL & 14K isolated videos, 9K words/phrases \\
        ASLDict & ASL & 48K isolated videos, 24K words/phrases \\
        \bottomrule
        \end{tabular}
    }
    \caption{Overview of the dictionary resources used in this work.}
    \label{tab:dictionaries}
\end{table}

\newpara{Architecture.}
We instantiate two model variants, one operating on RGB videos and one on keypoint sequences. For the keypoint-based model, we initialise from the pretrained pose backbone of~\cite{jang2025lost}, consisting of an Adaptive Graph Convolutional Network (AGCN)~\cite{shi2019two} followed by a Conformer~\cite{gulati20_interspeech}. This backbone produces a 512-dimensional feature representation. For the RGB model, we initialise from the pretrained SignRep model~\cite{wong2025signrep}, which uses a Video-Hiera transformer architecture~\cite{ryali2023hiera} and produces a 768-dimensional feature representation.

For Stage~1, the backbone $b_\theta$ processes co-articulated sign segments and the projection head $\pi_\theta$ consists of a linear layer that maps backbone features to a shared 512-dimensional embedding space. For Stage~2, the dictionary projection network $g_\phi$ is implemented as a two-layer LSTM~\cite{hochreiter1997long} with hidden dimension 512. The prototype dimension is therefore fixed to $D=512$.

\newpara{Stage~1 training.}
Stage~1 is trained using temporally annotated sign segments from BSLCorpus~\cite{schembri2013building}. 
For each annotated instance, we extract a 32-frame temporal window centred on the target sign and use the associated \textit{id-gloss} label as supervision for learning the sign prototypes and embedding space. To address class imbalance, we use a balanced sampler with batch size 16.

We fine-tune the backbone, projection head, and prototype set end-to-end during Stage~1 training. All prototypes are randomly initialised in a 512-dimensional space. We set the prototype temperature using $\tau = \frac{1}{\log(K-1) - \log(1/\alpha - 1)}$
with $\alpha=0.9$, so that the resulting softmax initially assigns approximately 0.9 probability mass to the class associated with the input.
For each model variant, we use the same augmentation settings as those used for the corresponding backbone pretraining \cite{jang2025lost,wong2025signrep}.

We use AdamW \cite{loshchilovdecoupled} with learning rate $10^{-4}$, weight decay $10^{-4}$, and $\beta=(0.9,0.999)$ to train Stage~1. Models are trained for 25{,}000 steps with batch size 16 and gradient clipping at 1.0. The learning rate follows a warmup cosine schedule with 1{,}000 warmup steps and cosine decay thereafter.

\newpara{Stage~2 training.}
Stage~2 is trained using isolated dictionary videos from BSL SignBank~\cite{cormier2012corpus}, where each example is labelled with an \textit{id-gloss} corresponding to a Stage~1 sign class. 
Each dictionary video is converted into a sequence of 32-frame temporal windows extracted using a sliding window with stride 16 over the active signing interval. These windows are then encoded independently by the frozen backbone and used as input to the dictionary projection network.
For both pose and RGB variants, we use the same augmentation strategy as in Stage~1.

During Stage~2, the backbone and prototypes are initialised from the final Stage~1 checkpoint and frozen, while only the dictionary projection network is trained using the same optimiser and learning-rate schedule as Stage~1, but with batch size 64. We reuse the Stage~1 temperature $\tau$ and optimise $\mathcal{L}_{\mathrm{stage2}}$.

\newpara{Pseudo-label training on larger corpora.}
We refer to the supervised training procedure described above, consisting of Stage~1 and Stage~2 training on BSLCorpus and BSL SignBank, as \textbf{Phase~1}. To extend supervision beyond the manually annotated BSLCorpus, we next apply the trained model to large unlabelled continuous sign corpora, including YouTube-SL-25~\cite{tanzer2025youtube}, its BSL and ASL subsets, and BOBSL~\cite{Albanie2021bobsl}.

Using the corresponding dictionary resources, BSL SignBank and BSLDict for BSL, and ASLDict for ASL, we generate pseudo-labels using the sliding-window matching procedure described in \Cref{sec:dict_spotting_in_coart}. Additional details of the pseudo-label generation process are provided in the supplementary material.

These pseudo-labels are then used as additional supervision in \textbf{Phase~2}. \Cref{tab:data_growth} summarises the scale of the resulting training data. The model is retrained using the same architecture and optimisation settings as in Phase~1, except that training is extended to 125{,}000 optimisation steps to accommodate the substantially larger pseudo-labelled corpus.

\begin{table}[t]
    \centering
    \setlength{\tabcolsep}{20pt}
    \resizebox{\linewidth}{!}
    {
        \begin{tabular}{lrc}
        \toprule
        Training Setup & Training Samples & Matched Dictionary Signs \\
        \midrule
        \textbf{Phase~1} & 36K & 1.7K \\
        \midrule
        \textbf{Phase~2 / Pseudo-label Training} & & \\
        \quad BOBSL & 2,194K & 4.5K \\
        \quad YouTube-ASL & 957K & 5.6K \\
        \quad YouTube-BSL & 119K & 1.6K \\
        \quad BSLCorpus (manual annotations) & 36K & 1.7K \\
        \bottomrule
        \end{tabular}
    }
    \caption{
    Scale of the training data used across phases. Phase~2 expands supervision through pseudo-labels generated on large continuous sign language corpora using the corresponding dictionary resources.
    }
    \label{tab:data_growth}
\end{table}

\newpara{From BSL to ASL.}
To evaluate whether the learned representation transfers beyond the source language, we first train a model using only BSL supervision, consisting of BSLCorpus annotations and BSL SignBank dictionaries in \textbf{Phase~1}. We then expand training in \textbf{Phase~2} using pseudo-labels generated on a larger multilingual corpus, including YouTube-ASL, YouTube-BSL, BSLCorpus, and BOBSL, together with the corresponding dictionary resources. As shown in \Cref{tab:iterative_training}, this iterative procedure consistently improves dictionary retrieval performance on ASL-Citizen, demonstrating that a representation learned from BSL supervision can be effectively extended to an unseen sign language through the proposed sign matching framework. Phase~1 uses no pseudo-labels and already improves over the backbone features by 3.75 DCG, while Phase~2 provides further gains through the additional scale of pseudo-labels.

\begin{table}[ht]
\centering
\small
\begin{tabular}{lccc}
\toprule
\textbf{Training Stage} & \textbf{DCG} & \textbf{R@1} & \textbf{R@5} \\
\midrule
Pretrained SignRep (SSL) \cite{wong2025signrep} 
& 71.21 & 49.95 & 80.09 \\
After Phase 1 training
& 74.96 & 54.92 & 84.18 \\
After Phase 2 training
& \textbf{83.17} & \textbf{67.12} & \textbf{91.78} \\
\bottomrule
\end{tabular}

\caption{
Effect of iterative training on ASL-Citizen dictionary retrieval performance using SignRep as the RGB backbone. Phase 1 is trained using BSLCorpus annotations and BSL SignBank dictionary supervision. Phase 2 extends training with pseudo-labelled data generated from large multilingual sign language corpora and their corresponding dictionary resources.
}
\label{tab:iterative_training}
\end{table}

%% file: secs/5_exps.tex
\section{Applications and Generalisation}

We evaluate our matching network on three tasks that assess the transferability and scalability of the learned representation.
Unless stated otherwise, we use the frozen model obtained after 
Phase~2 training, and evaluate it zero-shot on all tasks.

(i) \textit{Dictionary-to-dictionary retrieval.} We evaluate retrieval on ASL-Citizen \cite{desai2023asl}, a crowd-sourced webcam dataset containing 2,731 isolated signs captured across diverse participants and recording conditions. This task measures whether the learned representation generalises to unseen signs language dataset. Performance is reported using Discounted Cumulative Gain (DCG), Recall@1, and Recall@5.

(ii) \textit{Dictionary-based sign matching} on the ChaLearn OSLWL Sign Spotting Challenge \cite{vazquez2022eccv}. As the benchmark is Spanish Sign Language, unobserved during training, this is a fully cross-lingual evaluation. We report the official average F1.

(iii) \textit{Automatic data labelling via dictionary spotting} on BOBSL against CSLR2 reference annotations \cite{raude2024}, comparing with both dictionary- and subtitle-based annotation approaches. We report WER, mIoU and mean F1 at multiple IoU thresholds.

\subsection{Dictionary-to-Dictionary Retrieval}

We evaluate dictionary-to-dictionary retrieval on ASL-Citizen without any training or fine-tuning on isolated sign videos from the dataset. Retrieval is performed using normalised dictionary embeddings produced by $g_\phi$, following the evaluation protocol of ASL-Citizen~\cite{desai2023asl}.

As shown in \Cref{tab:retrieval_aslcitizen_d2d}, our model achieves strong performance in terms of DCG, Recall@1, and Recall@5. Prior work demonstrated that strong retrieval performance typically requires direct supervision on ASL-Citizen \cite{desai2023asl}, as reflected by the substantial performance gap between I3D models trained on Kinetics \cite{carreira2017quo}, WLASL \cite{li2020word}, and ASL-Citizen itself. In contrast, our RGB model achieves state-of-the-art retrieval performance without any training on ASL-Citizen, surpassing even models trained directly on the target dataset. 
Notably, the pretrained SignRep backbone achieves 71.21 DCG, below the 76.52 DCG of I3D trained directly on ASL-Citizen. Starting from the same pretrained backbone, our two-stage training increases performance to 83.17 DCG, an improvement of 11.96 DCG over SignRep and 6.65 DCG over the ASL-Citizen trained baseline. These results suggest that the learned dictionary representation transfers effectively across datasets, enabling robust retrieval without dataset-specific supervision.

\begin{table}[th]
\centering
\small
\begin{tabular}{lccc}
\toprule
\textbf{Features} & \textbf{DCG} & \textbf{R@1} & \textbf{R@5} \\
\midrule
% \midrule
I3D Features (Kinetics) $^*$ \cite{carreira2017quo} & 12.34 & 0.41 & 1.36 \\
I3D Features (WLASL) $^*$ \cite{li2020word} & 31.53 & 9.85 & 25.20 \\
SignRep (SSL) \cite{wong2025signrep} & 71.21 & 49.95 & 80.09 \\
I3D Features (ASL-Citizen) \cite{desai2023asl}& 76.52 & 58.03 & 84.50 \\
\midrule
\textbf{Ours (Pose)} & 75.88 & 57.32 & 83.67 \\
\textbf{Ours (RGB)} & \textbf{83.17} & \textbf{67.12} & \textbf{91.78} \\
\bottomrule
\end{tabular}
\caption{Dictionary-to-dictionary retrieval performance on ASL-Citizen. Despite not being trained on ASL-Citizen, our proposed RGB model achieves state-of-the-art retrieval performance across all metrics. $^*$ denotes results reported by \cite{desai2023asl}}
\label{tab:retrieval_aslcitizen_d2d}
\end{table}

\subsection{Dictionary-Based Sign Matching}

We evaluate dictionary-based sign matching on the ChaLearn One-Shot Learning and Weak Labels (OSLWL) benchmark \cite{vazquez2022eccv}. The task requires localising occurrences of isolated dictionary signs within continuous signing sequences using a single dictionary exemplar per query. Each query is represented using dictionary embeddings, and detections are obtained through similarity-based matching over sliding temporal windows, following the retrieval procedure described in \Cref{sec:dict_spotting_in_coart}.

As shown in \Cref{tab:spotting}, our approach achieves state-of-the-art performance under the official F1 metric, substantially outperforming both the baseline and the best reported competition result. These results demonstrate accurate temporal localisation and effective cross-lingual transfer, despite training without Spanish Sign Language supervision.

\begin{table}[t]
\centering
\small
\begin{tabular}{lc}
\toprule
\textbf{Method} & \textbf{avg F1} \\
\midrule
Baseline \cite{vazquez2022eccv} & 0.395 \\
\raisebox{0.5ex}{$\drsh$}Best Reported Competition Result& 0.596 \\
\midrule
\textbf{Ours (RGB)} & 0.684 \\
\textbf{Ours (Pose)} & \textbf{0.695} \\
\bottomrule
\end{tabular}
\caption{Dictionary-to-co-articulated sign matching results on the ChaLearn OSLWL benchmark. Our method achieves state-of-the-art performance under the official average F1 metric.}
\label{tab:spotting}
\end{table}

\subsection{Automatic Data Labelling via Dictionary Spotting}

We evaluate our approach as an automatic annotation method using the CSLR2~\cite{raude2024} evaluation protocol on the BOBSL dataset. This provides a direct comparison between scalable approaches for automatic sign language annotation over large-scale datasets in an open-vocabulary setting. Unlike prior approaches that combine dictionary supervision with subtitle alignment, mouthings, or translation-based supervision, our framework performs labelling directly through dictionary-based visual retrieval.

As shown in \Cref{tab:labeling}, the proposed method significantly improves over prior automatic annotation approaches across WER, mIoU, and F1 metrics. Previous approaches combine multiple supervisory signals, including mouthings (M), dictionary supervision (D), and attention-based alignment from sign language translation models (A). In contrast, our annotation pipeline does not require mouthings, subtitle alignment, or translation-based supervision, yet achieves stronger performance.

We further compare against the dictionary-based sign spotting framework of \cite{momeni2020watch}, which combines sparse annotations, subtitles, and dictionary supervision using MIL and NCE. In contrast, our simpler retrieval approach uses learned dictionary embeddings and sliding-window matching, without subtitle-driven optimisation or multi-stage supervision, while achieving stronger transfer from isolated to co-articulated signing across all metrics.

An important observation from \Cref{tab:labeling} is that the performance of our method changes only marginally when subtitle-based dictionary filtering is removed, whereas prior approaches rely heavily on subtitle information. In the subtitle setting, retrieval is restricted to dictionary entries whose original, text normalised, or lemmatised forms match words appearing in the accompanying subtitle text. Retrieval is then performed only over the resulting subset of candidate dictionary entries. In the no-subtitle setting, retrieval is performed over the full dictionary vocabulary. We hypothesise that subtitle supervision provides only an indirect proxy for the signed content, since subtitles may be loosely aligned with the video and do not necessarily correspond one-to-one with the signed expression. We provide further qualitative examples in the supplementary material.

\begin{table*}[t]
    \centering
    \setlength{\tabcolsep}{4pt}
    \resizebox{\linewidth}{!}
    {
        \begin{tabular}{lccccc}
        \toprule
        \textbf{Method} 
        & \textbf{WER $\downarrow$} 
        & \textbf{mIoU $\uparrow$} 
        & \textbf{F1@0.1 $\uparrow$} 
        & \textbf{F1@0.25 $\uparrow$} 
        & \textbf{F1@0.5 $\uparrow$} \\
        \midrule
        Subtitle-based auto. annots. $[\text{M}_{.5}\text{D}_{.7}\text{A}_{.0}]$ \cite{Albanie2021bobsl}
        & 115.8 & 13.0 & - & - & -   \\
        Subtitle-based auto. annots. $[\text{M}_{.8}\text{D}_{.8}\text{A}]$ \cite{Albanie2021bobsl}
        & 93.8 & 13.0 & - & - & -   \\
        Subtitle-based auto. annots. $[\text{M}^*_{.8}\text{D}^*_{.8}\text{A}_{.0}\text{P}_{.5}\text{E }\text{N}]$  \cite{momeni2022automatic} 
        & 90.7 & 20.0  & - & - & -  \\
        Subtitle-based auto. annots. $[\text{M}^*_{.8}\text{D}^*_{.8}\text{A}_{.0}\text{P}_{.5}]$  \cite{momeni2022automatic} 
        & 85.5  & 18.1 & - & - & - \\
        Subtitle-based auto. annots. $[\text{M}^*_{.7}\text{D}^*_{.9}\text{A}_{.4}\text{P}_{.2}]$ \cite{momeni2022automatic} 
        & 85.9  & 15.3 & - & - & - \\
        \midrule
        Dictionaries \cite{momeni2020watch}
        & 91.81 & 9.24 & 13.09 & 12.41 & 7.56 \\
        Ours (RGB) (using subtitle) 
        & 80.93 & 20.05 & 27.66 & 27.18 & 23.41 \\
        Ours (Pose) (using subtitle) 
        & 82.54 & 17.79 & 25.05 & 24.71 & 21.68 \\
        Ours (RGB) (no subtitle) 
        & 81.05 & 21.18 & 28.80 & 28.45 & 25.34  \\
        Ours (Pose) (no subtitle) 
        & \textbf{79.57} & \textbf{21.66} & \textbf{29.48} & \textbf{29.25} & \textbf{26.30} \\
        \bottomrule
        \end{tabular}
    }
    \caption{Automatic data labelling performance on the BOBSL test set evaluated against CSLR2 annotations. We report WER, mIoU, and mean F1 at multiple IoU thresholds.}
    \label{tab:labeling}
\end{table*}

\subsection{Ablation Studies}

\paragraph{Impact of the two-stage pipeline.}

As an additional ablation, we replace the proposed two-stage framework with a standard pose-based classification model while keeping the same pose backbone architecture. The model is trained jointly on dictionary and co-articulated signing data using a cross-entropy objective over the shared dictionary vocabulary. Dictionary videos are uniformly sampled to 32 frames while preserving coverage of the full isolated sign sequence. During inference, we extract global average pooled features and perform retrieval using the same sliding-window matching procedure as our method.

As shown in \Cref{tab:ab1}, this classification-based baseline relies more heavily on subtitle information, and its performance degrades substantially when subtitles are removed. In contrast, our approach remains considerably more robust in the no-subtitle setting and achieves higher overall retrieval performance. These results suggest that directly optimising a closed-vocabulary classification objective does not produce representations that transfer as effectively to open-vocabulary sign retrieval.
We further ablate the training objective and analyse its effect on the learned prototype structure in the supplementary material.

\begin{table*}[t]
    \centering
    \setlength{\tabcolsep}{20pt}
    \resizebox{\linewidth}{!}
    {
        \begin{tabular}{lcccc}
        \toprule
        & \multicolumn{2}{c}{\textbf{With Subtitle}} 
        & \multicolumn{2}{c}{\textbf{No Subtitle}} \\
        \cmidrule(lr){2-3} \cmidrule(lr){4-5}
        \textbf{Method} 
        & \textbf{WER $\downarrow$} 
        & \textbf{mIoU $\uparrow$}
        & \textbf{WER $\downarrow$} 
        & \textbf{mIoU $\uparrow$} \\
        \midrule
        Classification Model 
        & 83.80 & 16.83 
        & 88.18 & 11.70 \\
        Ours
        & \textbf{82.54} & \textbf{17.79} 
        & \textbf{79.57} & \textbf{21.66} \\
        \bottomrule
        \end{tabular}
    }
    \caption{
    Impact of the two-stage pipeline on automatic data labelling performance on CSLR2. Compared with a classification-based baseline, the proposed sign matching framework achieves higher retrieval performance and remains substantially more robust when subtitle supervision is removed.}
    \label{tab:ab1}
\end{table*}

\paragraph{Cross-lingual transfer to ASL.}
To evaluate cross-lingual transfer, we train the model using only BSL supervision, including BSL dictionary data and pseudo-labelled BSL corpora, and evaluate directly on ASL-Citizen without ASL-specific supervision. As shown in \Cref{tab:retrieval_aslcitizen}, the resulting model achieves strong retrieval performance and outperforms both original SignRep pretrained model and the ASL-Citizen I3D features. Incorporating ASL data during training provides a further improvement, increasing DCG from 80.35 to 83.17 and Recall@1 from 62.75 to 67.12. These results show that the representation transfers across sign languages without language-specific supervision, while also benefiting from increased linguistic diversity during training.

\begin{table}[th]
\centering
\small
\begin{tabular}{lccc}
\toprule
\textbf{Features} & \textbf{DCG} & \textbf{R@1} & \textbf{R@5} \\
\midrule
SignRep (SSL) \cite{wong2025signrep} & 71.21 & 49.95 & 80.09 \\
I3D Features (ASL-Citizen) \cite{desai2023asl} & 76.52 & 58.03 & 84.5 \\
\midrule
Ours (RGB) - Trained on BSL only & {80.35} & {62.75} & {89.47} \\

Ours (RGB) - Trained on ASL+BSL & \textbf{83.17} & \textbf{67.12} & \textbf{91.78} \\
\bottomrule
\end{tabular}
\caption{Dictionary-to-dictionary retrieval performance on ASL-Citizen. A model trained using only BSL supervision already outperforms prior ASL-specific baselines, while incorporating both BSL and ASL data provides further gains.}
\label{tab:retrieval_aslcitizen}
\end{table}

\subsection{Qualitative analysis}

\begin{figure}[ht]
    \centering
    \includegraphics[width=0.9\linewidth]{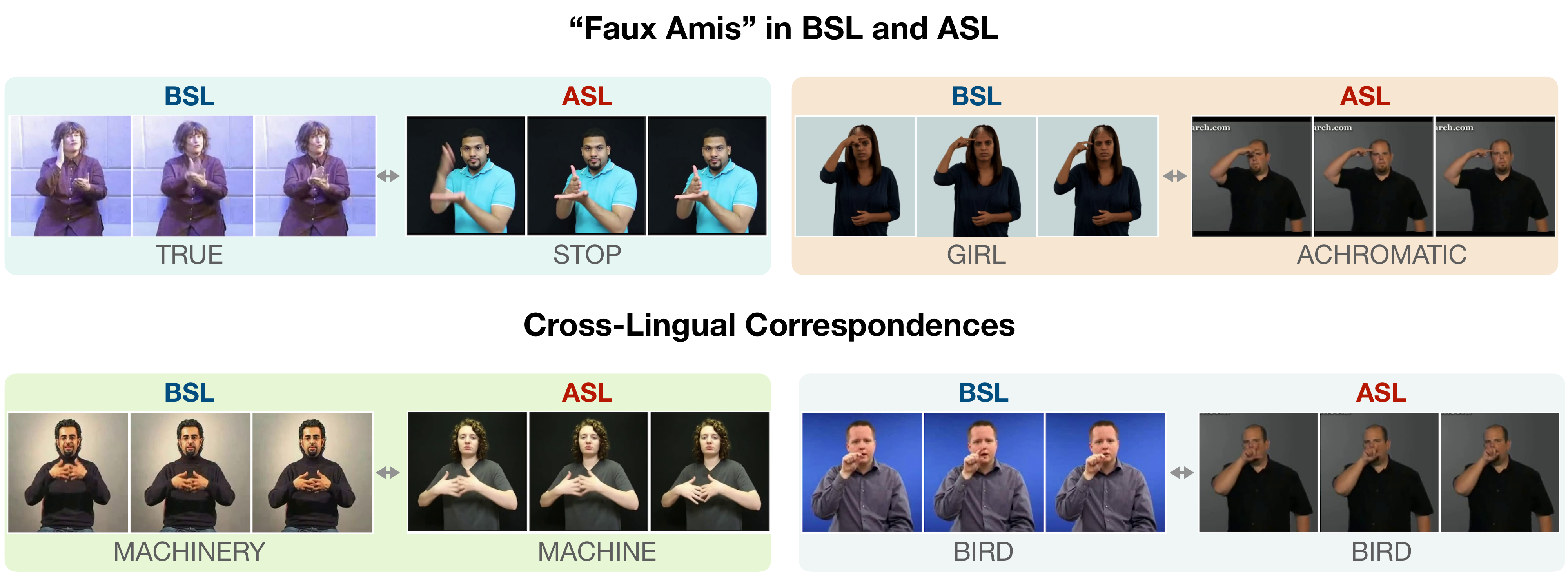}
    \caption{
    \textbf{Cross-lingual lexical similarities and potential faux amis.} Examples retrieved across BSL and ASL using learned dictionary embeddings. The top row shows visually similar signs with different meanings (potential faux amis), while the bottom row shows signs with related meanings, illustrating the learned cross-lingual structure.
    }
    \label{fig:faux}
\end{figure}

\newpara{Cross-lingual lexical similarities and potential faux amis.}
Although the training objective does not explicitly model cross-lingual relationships, the learned dictionary embeddings retrieve visually similar signs across BSL and ASL, as shown in \Cref{fig:faux}. The top row shows visually similar signs with different meanings (potential faux amis), whereas the bottom row shows signs with related or equivalent meanings across languages. These results suggest that our model may support  large-scale studies of lexical similarities across sign languages \cite{mckee2000lexical,aldersson2007lexical}.

%% file: secs/6_conclusion.tex
\section{Conclusions}

We introduced sign matching, a purely visual task that localises dictionary signs in continuous co-articulated signing video without relying on an intermediate textual representation. Our approach decouples representation learning from dictionary alignment by first learningan embedding space organised by visual similarity from continuous sign corpora and then aligning isolated dictionary exemplars to that space. This enables the model to exploit the natural variation present in continuous signing while adapting to dictionary resources that contain only one or a few examples per sign.

Across ASL-Citizen dictionary retrieval, ChaLearn OSLWL sign matching, and BOBSL automatic annotation evaluated using CSLR2 labels, we showed that the learned representation transfers beyond the initial supervised vocabulary to new datasets, tasks, and sign languages. The model generalises from BSL and ASL supervision to Spanish Sign Language sign matching, while also supporting large-scale automatic annotation through dictionary-guided retrieval. Iterative pseudo-labelling on weakly aligned corpora further improves performance and enables scaling to larger multilingual collections. These results suggest that dictionary-driven sign matching can provide a practical bridge between isolated sign resources and continuous sign language corpora, enabling sign-centric video search, corpus navigation, and large-scale annotation, and taking a step towards a universal sign representation that transfers across sign languages.

%% file: secs/7_appendices.tex
\section*{SignMatch - Supplementary Material}

We describe the pseudo-labelling pipeline used to scale training to large continuous sign language corpora in Sec.~A. Qualitative examples of dictionary matching within continuous signing, together with an analysis of subtitle-guided dictionary filtering, are given in Sec.~B. Sec.~C evaluates soft versus hard training targets and analyses their effect on the learned prototype space. Finally, we discuss limitations and future directions in Sec.~D.

\subsection*{A. Scaling Sign Matching for Automatic Sign Annotation}
\label{app:scaling}

We begin with a small set of manually aligned sign instances from BSLCorpus, where co-articulated sign segments are annotated with BSL SignBank dictionary identifiers and temporal boundaries. These paired dictionary and continuous signing examples provide the initial supervision required to learn a shared embedding space through the two-stage training procedure described in the main paper.

Starting from this initial supervision, we scale training by automatically discovering additional sign instances in large-scale continuous sign language corpora. Rather than relying exclusively on manually annotated examples, we leverage subtitle-aligned videos from BOBSL and the BSL and ASL portions of YouTube-SL-25 together with dictionary entries from BSL SignBank (BSL), BSLDict (BSL), and ASLDict (ASL). The learned sign matching model is used to retrieve candidate occurrences of dictionary signs within continuous signing data, enabling the automatic construction of substantially larger training sets. The process consists of the following stages:

\newpara{Caching Dictionary Embeddings.}
We first apply the dictionary branch of the model to all dictionary videos and cache the resulting dictionary embeddings, $e^d$, together with their associated dictionary identifiers and keywords. This one-time computation enables efficient large-scale retrieval without repeatedly processing the dictionary videos.

\newpara{Query Selection.}
To improve the quality of the pseudo-labels used for Phase~2 training, we use subtitle text to constrain the set of dictionary queries. Given a subtitle segment, we first lemmatize the text and retain only words that match dictionary keywords from the corresponding sign language dictionaries. Each retained keyword is then mapped to one or more dictionary embeddings, producing a candidate query set for retrieval. This vocabulary filtering reduces the likelihood of spurious matches when applying a model trained on the comparatively small BSLCorpus dataset to large-scale continuous sign language corpora, thereby reducing false positive pseudo-labels and providing more reliable supervision for retraining.

\newpara{Sliding-Window Retrieval.}
For each subtitle-aligned video clip, we densely sample temporal windows from the continuous signing sequence using a sliding-window strategy with a stride of 2. 
The co-articulated embedding for each window, $e$, is obtained from the output of $\pi_\theta$. Retrieval is performed by comparing each window embedding against the candidate dictionary embeddings identified during query selection using cosine similarity. Windows whose similarity score exceeds a threshold of 0.5 are retained as candidate sign detections.

\newpara{Post-processing.}
To improve label quality, we apply a series of filtering steps. First, temporally overlapping detections with the same label are merged by aggregating contiguous detections and averaging their scores. We additionally apply overlap-based suppression to remove duplicate detections, retaining only the highest-scoring segment when the temporal intersection-over-union exceeds a threshold of 0.5.

\newpara{Output.}
The resulting pseudo-labelled dataset consists of temporally localised sign segments paired with dictionary labels and confidence scores. These annotations substantially expand the available supervision and are combined with the original manually annotated training data. Together with the corresponding dictionary videos from BSL SignBank, BSLDict, and ASLDict, they are used as supervision during Phase~2 retraining.

\begin{figure}[ht]
    \centering
    \includegraphics[width=0.9\linewidth]{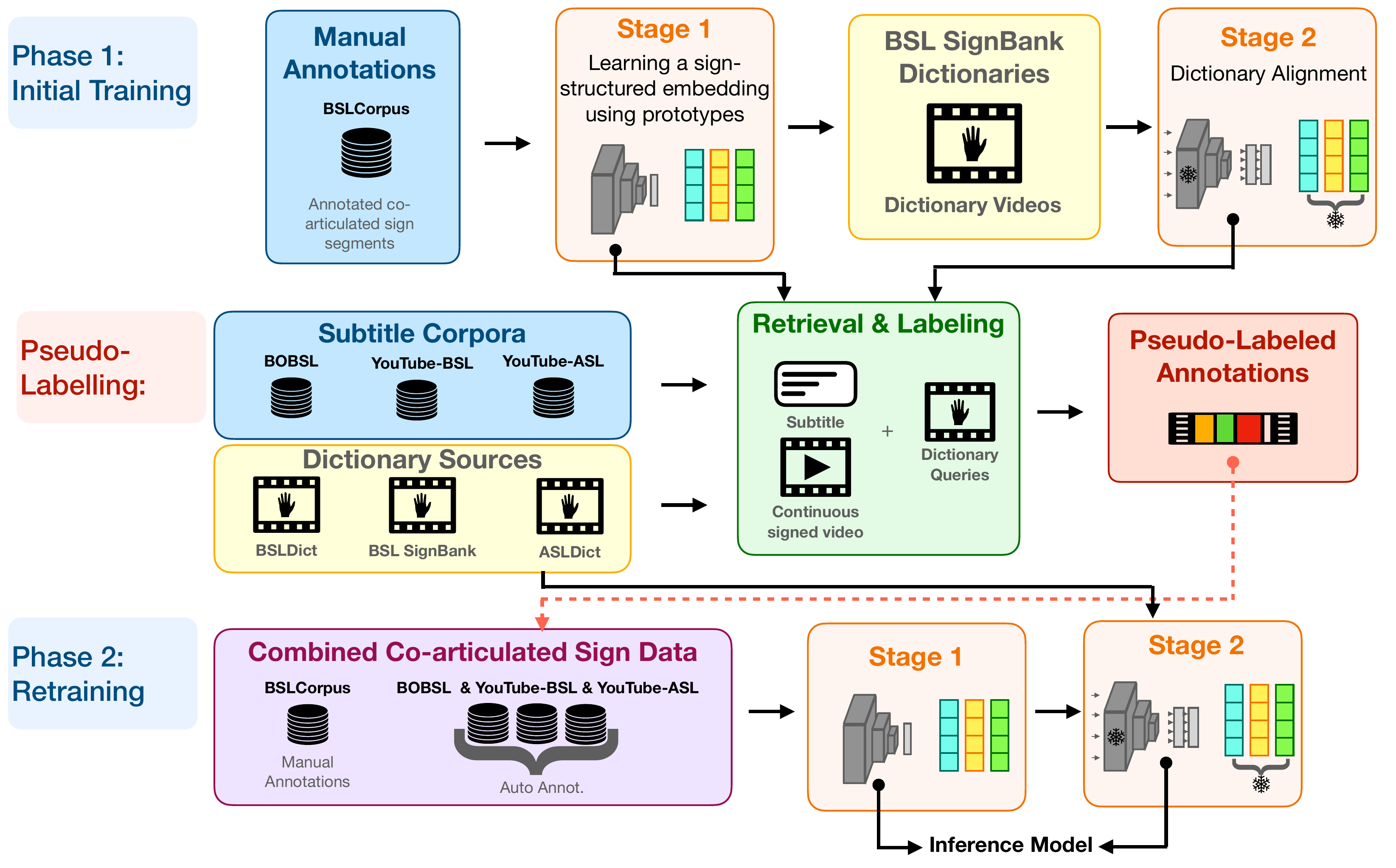}
    \caption{Overview of the proposed training and pseudo-labelling pipeline. A sign matching model is first trained using manually annotated BSLCorpus sign segments and SignBank dictionary videos. The learned representation is then used to automatically retrieve and temporally localise sign occurrences in large-scale subtitle-aligned sign language corpora using dictionary queries from BSL SignBank, BSLDict, and ASLDict. The resulting pseudo-labelled annotations are combined with the original manual annotations and used for Phase 2 retraining, producing a more robust sign matching model.}
    \label{fig:fullpipeline}
\end{figure}

\Cref{fig:fullpipeline} summarises the full training, pseudo-labelling, and retraining procedure used in our approach.

\subsection*{B. Qualitative Analysis}
\label{app:qual}
\begin{figure}[t]
\centering
\includegraphics[width=1.0\linewidth]{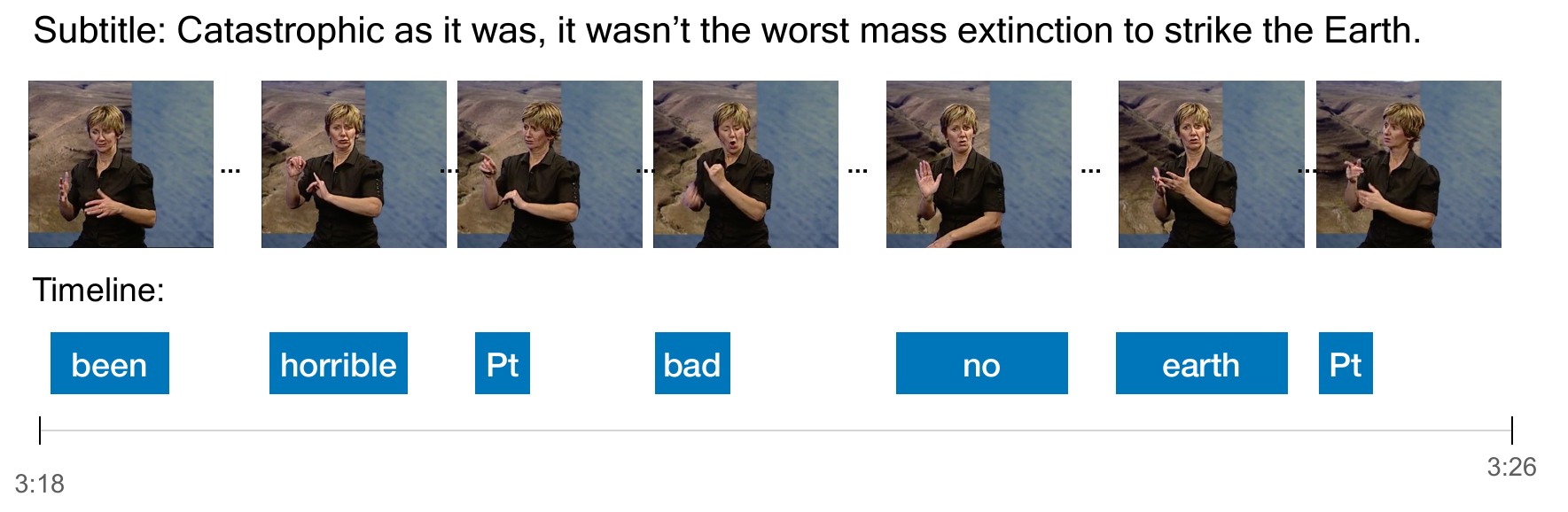}
\caption{Example dictionary label matches within a continuous sign language segment. Temporal intervals matched to dictionary labels are shown on the timeline, with a representative keyframe from each matched interval displayed above. The label \textit{Pt} corresponds to the dictionary gloss \textit{PT:PRO3SG}, a third-person pointing sign used to refer to an entity such as ``it’’ in the accompanying subtitle.}
\label{fig:timetovis}
\end{figure}

In \Cref{fig:timetovis}, we visualise dictionary label matches within a continuous sign language segment. Each interval on the timeline corresponds to a segment matched to a dictionary label, while the keyframes shown above are extracted from the associated matched segment and provide representative examples of the corresponding sign. The subtitle is included to provide linguistic context for the signed content.

\begin{figure}[t]
    \centering
    \includegraphics[width=1.0\linewidth]{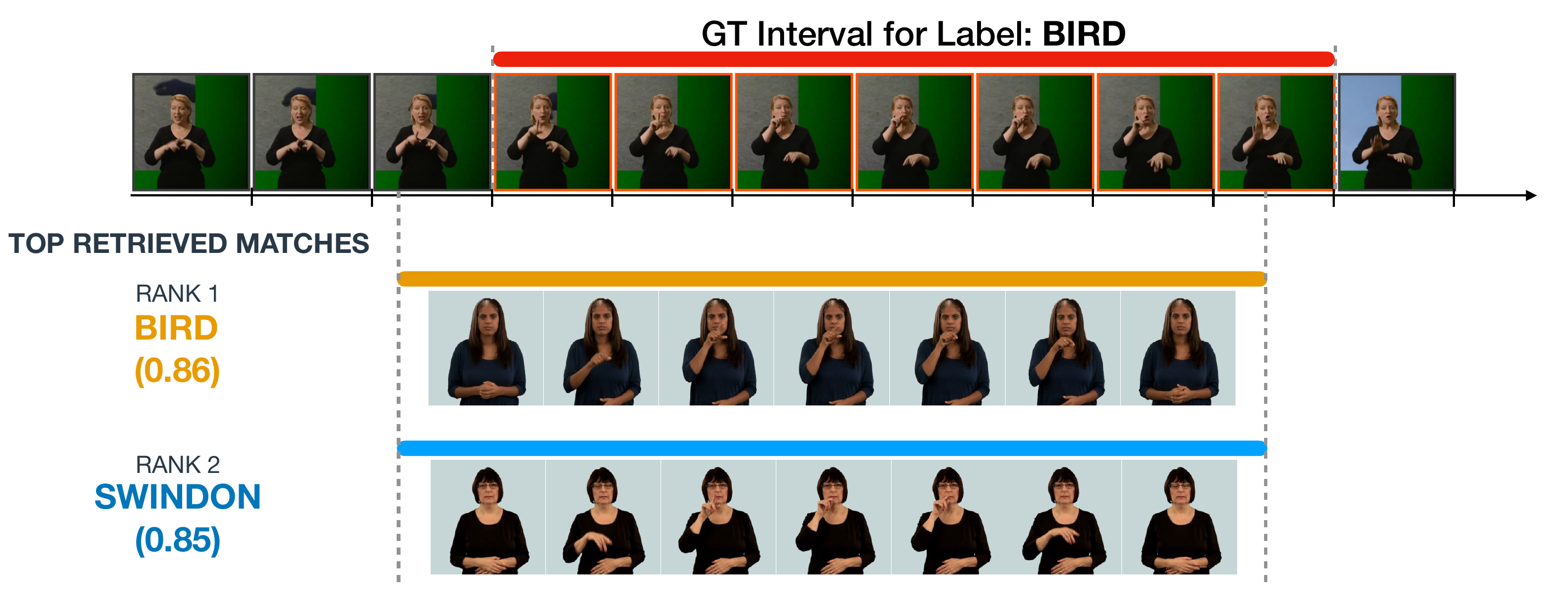}
    \caption{Qualitative retrieval results. The correct dictionary sign (\textit{BIRD}) is retrieved and localised within the continuous signing sequence. A visually similar sign (\textit{SWINDON}) also receives a high similarity score.}
    \label{fig:vis1}
\end{figure}

In \Cref{fig:vis1}, our model retrieves visually similar dictionary matches for a query sign and can surface multiple plausible candidates when they share similar production-level features. In this example, both \textit{BIRD} and \textit{SWINDON} are strongly matched to the same temporal region, reflecting their visual similarity despite representing different lexical items. The correct sign is ranked first, while the second ranking of \textit{SWINDON} illustrates that the representation places signs of similar visual production close together even when their meanings are unrelated. Incorporating additional cues, such as linguistic context, fingerspelling, and mouthing information, may further improve the ability to distinguish between visually similar signs and refine retrieval performance.

\subsubsection*{Subtitle Filtering for Automatic Annotation}

In \Cref{tab:qualitative}, we compare retrieval results obtained with and without subtitle-guided dictionary filtering on BOBSL using the CSLR2 evaluation protocol, highlighting the effects of subtitle-guided dictionary filtering on retrieval performance. The examples illustrate that subtitle filtering can have markedly different effects depending on the instance. In many cases, restricting retrieval to subtitle-derived dictionary entries removes valid sign detections and reduces recall. In other cases, subtitle information can suppress false positive matches or help resolve ambiguities that are difficult to distinguish from visual evidence alone. We also observe examples where both approaches produce similar predictions, indicating that the retrieval model itself is often the dominant factor.

Several factors contribute to these behaviours. First, subtitles may be temporally misaligned with the signed content, causing relevant signs to occur outside the corresponding subtitle segment. Second, subtitles do not necessarily provide a one-to-one lexical transcription of the signed utterance. Signers may omit words, introduce additional concepts, or express the same meaning using different lexical choices. Third, subtitle text reflects the grammar and vocabulary of the spoken language rather than the linguistic structure of the signed language. As a result, subtitle-derived keywords may exclude valid dictionary candidates, but can also act as a useful prior for suppressing some false positive retrievals.

The examples in \Cref{tab:qualitative} reflect this trade-off. Examples 1–4 show cases where subtitle filtering suppresses correct detections and reduces annotation coverage. Example 5 demonstrates a case where subtitle constraints help recover a more accurate prediction. Example 6 highlights a more subtle benefit: the glosses \textit{month} and \textit{minute} share a highly similar visual form and are often difficult to distinguish from visual evidence alone. In this case, subtitle filtering suppresses a plausible but incorrect match (\textit{minute}), illustrating how subtitle information can reduce false positives during automatic label generation. Example 7 illustrates another nuanced case: subtitle filtering removes a spurious prediction (\textit{stone}), producing a cleaner output, but does not change the final CSLR2 WER because the remaining matched glosses are unchanged.

Overall, the examples in \Cref{tab:qualitative} help explain the behaviour observed in \Cref{tab:labeling}. Subtitle-guided filtering can suppress some false positive matches and help resolve ambiguities that are difficult to distinguish from visual evidence alone, as illustrated by Examples 6 and 7. However, the examples also show that subtitle constraints can remove valid dictionary candidates and suppress correct detections, as in Examples 1--4. Taken together, these observations support the hypothesis that subtitle information provides only an indirect proxy for the signed content: it can be useful in specific cases, but does not consistently improve retrieval quality and may reduce annotation coverage.

\begin{table*}[htbp]
    \centering
    \footnotesize
    \setlength{\tabcolsep}{8pt}
    \renewcommand{\arraystretch}{1.2}
    \resizebox{\textwidth}{!}{
\begin{tabular}{@{}c p{2.3cm} p{2.7cm} p{3.2cm} p{2.6cm}@{}}
\toprule
        No. & Subtitle & GT & Pred (no subtitle) & Pred (with subtitle) \\
        \midrule

        1 & Lions are all over the place. &
        dangerous &
        \ok{danger} &
        \miss{[]} \\ \rowsep

        2 & It just went on and on and on. &
        many &
        \ok{many} &
        \miss{[]} \\ \rowsep

        3 & One man believed we could. &
        need | beat | want | one &
        \ok{want} | \ok{beat} | \ok{want} | \ok{one} &
        \ok{one} \\ \rowsep

        4 & They've brought alarming news. &
        happen | news | shock/alarm &
        \ok{happen} | \ok{news} | \ok{surprise} &
        \ok{news} \\ \rowsep

        5 & Angus cows to sell. &
        cow | sell | small | cow &
        \ok{cow} | \miss{pay} | \miss{reduce} | \ok{cow} &
        \ok{cow} | \ok{sell} | \ok{cow} \\ \rowsep

        6 & 18 months to two years. &
        month | two | year &
        \miss{minute} | \ok{year} | \miss{wherever} &
        \ok{month} | \ok{year} \\ \rowsep

        7 & What happened to the tomato farmers? &
        tomato | farmer | what &
        \miss{stone} | \ok{farm} | \ok{what} &
        \ok{farm} | \ok{what} \\

        \bottomrule
    \end{tabular}
    }
    \caption{\textbf{Qualitative comparison of gloss predictions.} Predicted glosses without subtitle-based dictionary filtering (\textit{Pred (no subtitle)}) and with it (\textit{Pred (with subtitle)}), against ground-truth glosses (GT). Following the CSLR2 evaluation protocol on BOBSL, matches include both exact word matches and words that are mapped to the same canonical label (e.g. synonyms and accepted lexical variants). \textcolor{matchgreen}{Green} marks matching tokens; \textcolor{missred}{red} marks mismatches or empty predictions (\texttt{[]}).}
    \label{tab:qualitative}
\end{table*}

\subsection*{C. Ablation: Training with Soft vs. Hard Targets}

To assess the benefit of soft targets within our two-stage training framework, we replace the prototype-induced soft targets with one-hot targets in both stages, using cross-entropy in place of the KL divergence objective while keeping the remaining training setup unchanged. We evaluate both objectives on BOBSL under the CSLR2 protocol, using the pose backbone without subtitle filtering. As shown in \Cref{tab:objective}, the soft-target objective performs better across all metrics, reducing WER from 82.25 to 79.57 and increasing mIoU from 17.48 to 21.66 and F1@0.1 from 24.93 to 29.48. These results demonstrate that preserving the similarity structure between sign classes through soft supervision contributes substantially to the performance of the two-stage framework.

\begin{table}[h]
\centering
\small
\begin{tabular}{lccc}
\toprule
\textbf{Objective} & \textbf{WER $\downarrow$} & \textbf{mIoU $\uparrow$} & \textbf{F1@0.1 $\uparrow$} \\
\midrule
Cross-entropy & 82.25 & 17.48 & 24.93 \\
Ours (KL) & \textbf{79.57} & \textbf{21.66} & \textbf{29.48} \\
\bottomrule
\end{tabular}
\caption{Ablation of soft versus hard targets within our two-stage training framework, evaluated on BOBSL under the CSLR2 protocol using the pose backbone without subtitle filtering. All other training settings are kept unchanged.}
\label{tab:objective}
\end{table}

\begin{figure}[ht]
    \centering
    \includegraphics[width=1.0\linewidth]{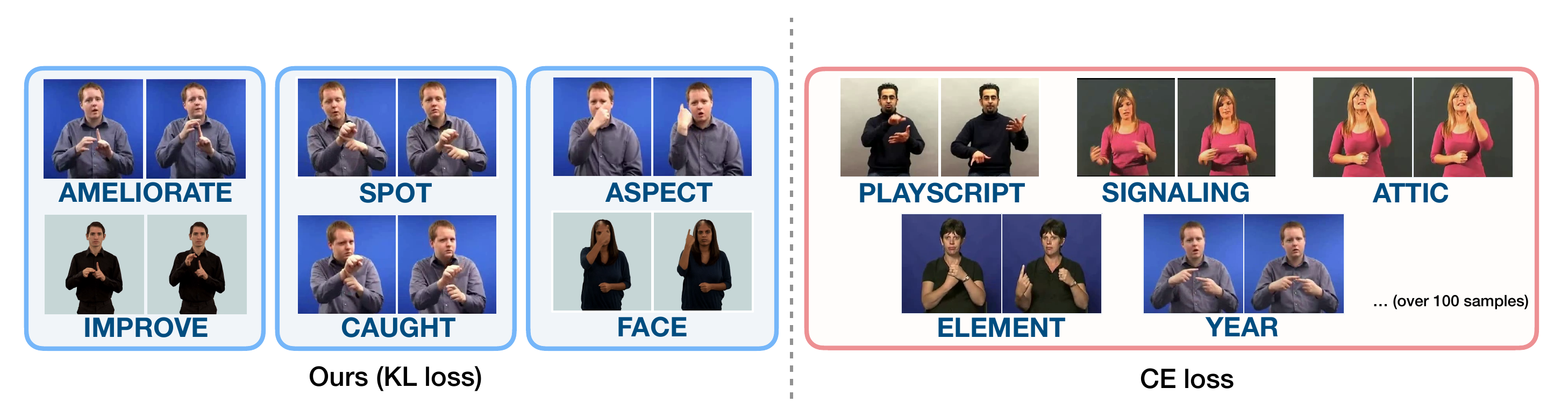}
    \caption{Comparison of the learned prototype space under soft-target and hard cross-entropy supervision at a cosine similarity threshold of $0.9$. Our soft-target objective forms small groups of visually similar sign classes, whereas cross-entropy groups prototypes from over one hundred sign classes into the same cluster.}
    \label{fig:clustering}
\end{figure}

\newpara{Prototype space analysis.}
To further examine how the training objective affects the learned representation, we cluster the prototypes at a cosine similarity threshold of $0.9$ for both the soft-target and hard cross-entropy models. As shown in \Cref{fig:clustering}, under cross-entropy, prototypes from over one hundred sign classes are grouped into the same cluster, despite substantial differences in their visual production. In contrast, our soft-target objective forms smaller groups of visually similar signs, such as \textit{SPOT} and \textit{CAUGHT}, \textit{ASPECT} and \textit{FACE}, and \textit{AMELIORATE} and \textit{IMPROVE}. This suggests that soft-target supervision encourages the prototype space to better reflect visual relationships between sign classes.

\subsection*{D. Limitations and Future Directions}

Our results indicate that the learned representation captures visual relationships between signs, as demonstrated by its transfer across datasets and sign languages and by the structure of the learned prototype space. A natural direction for future work is to further characterise this geometry in terms of specific production parameters, such as handshape, location, movement, and orientation, for example by comparing prototype neighbourhoods with phonological annotations. A complementary direction is to further reduce the gap between citation-form and co-articulated signing at the data level, for example by synthesising or augmenting faster variants of dictionary signs alongside the embedding-level alignment considered in this work.